\documentclass[10pt,twocolumn,letterpaper]{article}

\usepackage[pagenumbers]{cvpr} 

\usepackage{listings}
\usepackage{fvextra}

\definecolor{cvprblue}{rgb}{0.21,0.49,0.74}
\usepackage[pagebackref,breaklinks,colorlinks,allcolors=cvprblue]{hyperref}

\def\paperID{} 
\def\confName{CVPR}
\def\confYear{2025}

\newcommand{\ours}{SceneCast}
\newcommand{\trim}{ViTrim}  

\definecolor{turquoise}{cmyk}{0.65,0,0.1,0.3}
\definecolor{purple}{rgb}{0.65,0,0.65}
\definecolor{dark_green}{rgb}{0, 0.5, 0}
\definecolor{orange}{rgb}{0.8, 0.6, 0.2}
\definecolor{red}{rgb}{0.8, 0.2, 0.2}
\definecolor{darkred}{rgb}{0.6, 0.1, 0.05}
\definecolor{blueish}{rgb}{0.0, 0.3, .6}
\definecolor{light_gray}{rgb}{0.7, 0.7, .7}
\definecolor{pink}{rgb}{1, 0, 1}
\definecolor{greyblue}{rgb}{0.25, 0.25, 1}

\definecolor{thirdbestcolor}{rgb}{1,1, 0.6}
\definecolor{secondbestcolor}{rgb}{1, 0.9, 0.6}
\definecolor{firstbestcolor}{rgb}{1, 0.6, 0.6}

\usepackage{pgfplots}
\usepackage{subcaption}
\usepackage{amsmath}
\usepackage{multirow}
\usepackage{amssymb}
\usepackage{booktabs}
\pgfplotsset{compat=1.18}

\title{\ours: Learning Indoor Scene Synthesis from Curated Synthetic Videos with LLM Reward}

\author{Ziyang Zong\\
Sun Yat-sen University\\
{\tt\small zongziyang23@gmail.com}
\and
Canming Xia\\
Sun Yat-sen University\\
{\tt\small a1355134694@163.com}
\and
Haoran Xu\\
Peking University\\
{\tt\small xuhr.sz@pku.edu.cn}
\and
Guang Tan\\
Sun Yat-sen University\\
{\tt\small tanguang@mail.sysu.edu.cn}
}

\begin{document}
\maketitle
\begin{abstract}
3D indoor scene synthesis requires generating objects to form a complete and coherent scene. Existing approaches rely heavily on manually annotated datasets, leading to high costs and limited generalization. We present \ours, an annotation-free framework that trains on large-scale, unlabeled videos, addressing two key challenges: (1) Videos provide only partial views of a scene, requiring the model to extrapolate unseen 3D regions, which is inherently error-prone; (2) Raw videos available online are uncurated, suffering from uncontrolled conditions and variable data quality. First, we design a Transformer-based model that predicts objects sequentially and is fine-tuned with reinforcement learning (RL), where an LLM acts as a spatial common-sense evaluator. The LLM provides reward signals through chain-of-thought reasoning and code-executable geometric checks to refine object configurations. Second, we introduce \trim, a data curation approach that generates training videos with controllable video generation models. Crucially, it retains only high-quality prefix clips via multi-criteria quality assessment with an entropy-based weighting strategy. Evaluated on curated synthetic video sets, our annotation-free approach outperforms state-of-the-art supervised methods in scene quality.
\end{abstract}
\section{Introduction}

3D indoor scene synthesis is a core form of world model, with broad applications in 3D computer vision, computer graphics, and embodied AI~\citep{driess2023palm, huang2023embodied, yang2024holodeck, yang2024physcene}. Given an initial (possibly empty) set of objects placed in a scene, the task requires generating additional objects to produce a complete, coherent scene. The backbone network can be further trained with language conditions to support text-to-scene generation.

Recent approaches, including rule-based parametric modeling~\citep{purkait2020sg}, deep generative frameworks~\citep{wu2024external, bokhovkin2025scenefactor}, and 3D layout prediction pipelines~\citep{fang2025ctrl, feng2025casagpt}, have addressed various aspects of the task. 
However, these methods critically depend on large-scale, high-quality annotated indoor datasets such as 3D-FRONT~\citep{fu20213d} and ScanNet~\citep{dai2017scannet}, whose acquisition and annotation are costly and difficult to scale. This reliance has become a central bottleneck for further progress.

An alternative approach is to use Internet data, such as web images and videos for training. Recent reconstruction methods (e.g., VGGT~\citep{wang2025vggt}, Depth Anything~\citep{lin2025depth}, Lyra~\citep{bahmani2026lyra}, SceneGen~\citep{meng2026scenegen}) and recognition models can extract rich semantic 3D information from such data, \textit{eliminating the need for task-specific human annotation}. Internet videos are particularly appealing: they are abundant and provide broader viewpoint coverage than images. However, training a 3D scene generation model with videos online introduces two major challenges.

First, videos typically provide only partial observations of an indoor environment, requiring the generation model to extrapolate unseen regions in 3D space. Without effective constraints, such extrapolation easily causes error accumulation, geometric inconsistency, and violation of physical plausibility. Second, Internet videos are inherently uncurated along multiple dimensions: camera trajectories are uncontrolled and often degenerate (e.g., near-planar motion, abrupt cuts), scene content is non-static, and photometric conditions vary widely. These factors collectively degrade data quality, and addressing them demands broad data collection and heavy post-processing, with little guarantee of data quality.

\begin{figure*}
  \centering
    \includegraphics[width=0.9\textwidth]{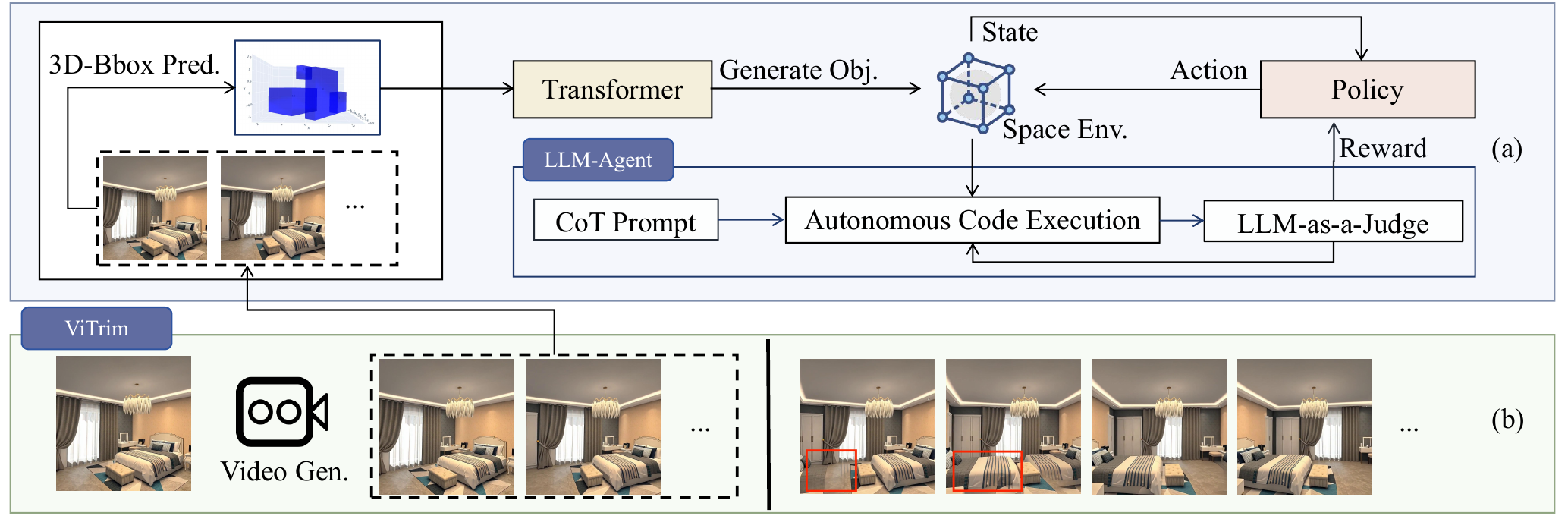}
    \caption{\ours\ overview. (a) Our generation model predicts 3D objects sequentially, starting from an initial set of objects. The model is fine-tuned with a reinforcement learning phase with LLM feedback. (b) Our \trim\ strategy removes low-quality tail segments from generated video using multiple quality assessment criteria, ensuring only valid clips are used for training.}
    \label{fig:intro}
\end{figure*}

To address the first challenge, we propose a Transformer model trained to predict 3D objects sequentially, then fine-tuned with reinforcement learning (RL). Each object is represented as a labeled 3D bounding box and instantiated with a 3D asset at render time. As shown in Figure~\ref{fig:intro}(a), given an initial set of objects, the model completes the scene by predicting the remaining objects one by one. To align the scene with spatial commonsense, RL iteratively refines the layout by adjusting object positions, sizes, and orientations, guided by reward signals from an LLM acting as a judge. The RL agent treats the current layout as its state and per-object attribute adjustments as actions. The LLM scores the layout against multiple criteria and returns a scalar reward to drive policy optimization. To improve scoring reliability, we prompt the LLM with Chain-of-Thought (CoT) reasoning backed by automatic code execution, enabling accurate geometric reasoning.

To address the quality control issue of raw Internet video, we propose a data curation approach called {\em 3D-Quality Aware Video Trimming}~(\trim). Rather than sourcing from the web, our approach generates training videos using recent video generation models. Pretrained on large-scale video data, these models encode strong priors over 3D object statistics and produce high-quality frames for short durations (e.g., a few seconds). Crucially, this approach offers control over scene content and camera trajectories, enabling targeted coverage of the indoor cases relevant to our task. A well-known problem with generated videos, however, is the error accumulation over long sequences. To address this, we propose a video trimming strategy, as shown in Figure~\ref{fig:intro}~(b). The method evaluates segments of each generated video, and trims the video to retain the initial high-quality clip. It evaluates multiple metrics, including human perceptual quality, spatial consistency, and physical plausibility, using a set of reward models~\citep{xu2023imagereward, xu2024visionreward, bansal2024videophy} whose scores are aggregated via {\em entropy-based weighting}. By doing so, our approach ensures that only usable data enters the 3D reconstruction pipeline.

Our main contributions are summarized as follows:
(1) An annotation-free 3D scene completion model that predicts 3D objects sequentially, and is fine-tuned with reinforcement learning using LLM feedback. 
(2) A data curation method that uses a video generation model to generate synthetic indoor scenes, followed by a quality-aware trimming strategy to retain only high-quality clips for training.
(3) A comprehensive performance evaluation on the 3D-FRONT benchmark that shows our model outperforms state-of-the-art methods.

\section{Related Work}

\textbf{Indoor Scene Generation.} Early methods for indoor scene generation use parametric methods~\citep{yu2015clutterpalette, fu2017adaptive, choi2015indoor} to explicitly model the spatial layout of indoor scenes. In contrast, data-driven deep learning approaches attempt to learn scene distributions from large-scale labeled data~\citep{wang2021sceneformer, li2018interiornet, zhong2025internscenes}. 
Recent research advances include autoregressive generation methods~\citep{ye2024maan, paschalidou2021atiss, chattopadhyay2023learning, choi2026scenenat, sun2024forest2seq}, latent space generation methods~\citep{purkait2020sg, lin2025scenelcm}, and diffusion generation methods~\citep{zhai2024echoscene, ju2024diffindscene, huang2025midi, tang2024diffuscene, meng2025lt3sd, bokhovkin2025scenefactor, chen2024instructscene}.
To reduce reliance on data, VIPScene~\citep{huang2025video} leverages video models to synthesize scenes with consistent views and broad viewpoint coverage, and concurrent works distill video diffusion models into explicit 3D representations~\citep{bahmani2026lyra, Zhang_2026_CVPR} or generate 3D scenes directly from a single image~\citep{meng2026scenegen, koo2026inspace}.
However, existing video and image driven methods fail to integrate indoor spatial common sense and global layout constraints into the generation process.

\textbf{LLM-Assisted Scene Generation.}
Prior works have explored using LLMs in indoor scene generation either as a semantic interface~\citep{yang2024holodeck, jain2023scenegpt, liu2023llm3dscene}, or as a prior encoder that provides design rules and spatial constraints~\citep{aguina2024open, zhang2024llmscene, yang2025optiscene, Jiang_2026_CVPR}.
Recently, LLMs have also been employed as reward evaluators, producing scalar reward signals or preference rankings to guide policy optimization~\citep{sun2025roomplanner, yuan2026planning, ran2025direct}.
Concurrent to our work, several agent-based systems~\citep{yang2026sceneweaver, pfaff2026scenesmith, xia2026sage, he2026sceneorchestra} employ LLM/VLM agents as explicit planners that iteratively construct scenes through tool calls.
However, these designs rely on the LLM's native inference, which lacks geometric reasoning accuracy. In contrast, our work uses the LLM as an agent that generates and invokes code for precise geometric computation, and optimizes a dedicated layout backbone rather than the language model itself. Our backbone is trained on video-derived layout statistics, providing a statistically grounded layout that agentic planners can refine with constraints.

\textbf{Generated Video Quality Evaluation.}
Existing video quality assessment methods operate along complementary dimensions: human-preference scores learned from annotated comparisons~\citep{he2024videoscore, jiang2023text2performer, shan2025populate, xu2024visionreward} capture perceptual fidelity, while VideoPhy~\citep{bansal2024videophy} evaluates compliance with physical commonsense. Benchmark suites such as VBench~\citep{huang2024vbench}, VBench2.0~\citep{zheng2025vbench}, and Video-Bench~\citep{han2025video} aggregate multiple such metrics for standardized model comparison.
However, none of these methods assess 3D spatial validity, which is the failure mode most critical to downstream 3D scene reconstruction. Moreover, multi-metric suites report scores independently without task-adaptive fusion.
Our \trim\ builds on VisionReward and VideoPhy as two of its evaluation dimensions, but introduces a spatial IoU signal targeting 3D structural quality, and fuses all dimensions via entropy-based weighting that adapts to each video's quality profile.

\section{Method}

In \ours, a 3D indoor scene is defined as an unordered set of 3D furniture objects $\mathcal{F} = \{f_v\}_{v=1}^N$. Each object $f_v$ has four attributes: $f_v = \left(c_v, p_v, u_v, d_v \right)$, where $p_v \in \mathbb{R}^3$ is the center position, $u_v \in \mathbb{R}$ is the rotation angle around the vertical (up) axis, $d_v \in \mathbb{R}^3$ is the object size, and $c_v \in \{1, 2, \dots, C\}$ denotes the object category.
The category vocabulary $\mathcal{C} = \{1, 2, \dots, C\}$ is defined by a predefined label set used during object recognition (via Grounded-SAM~\citep{ren2024grounded}) in the data preparation pipeline, and is therefore extensible: increasing the number of predefined labels directly increases $C$ without architectural changes. In 3D-FRONT experiments, we set $C=34$ to match the category of the baseline methods for fair comparison.
Given an initial (possibly empty) 3D objects set, we use a transformer to generate the remaining objects step by step. Commonsense knowledge is further integrated into the learning process through reinforcement learning that interfaces with an LLM.

\begin{figure*}
  \centering
    \includegraphics[width=0.85\textwidth]{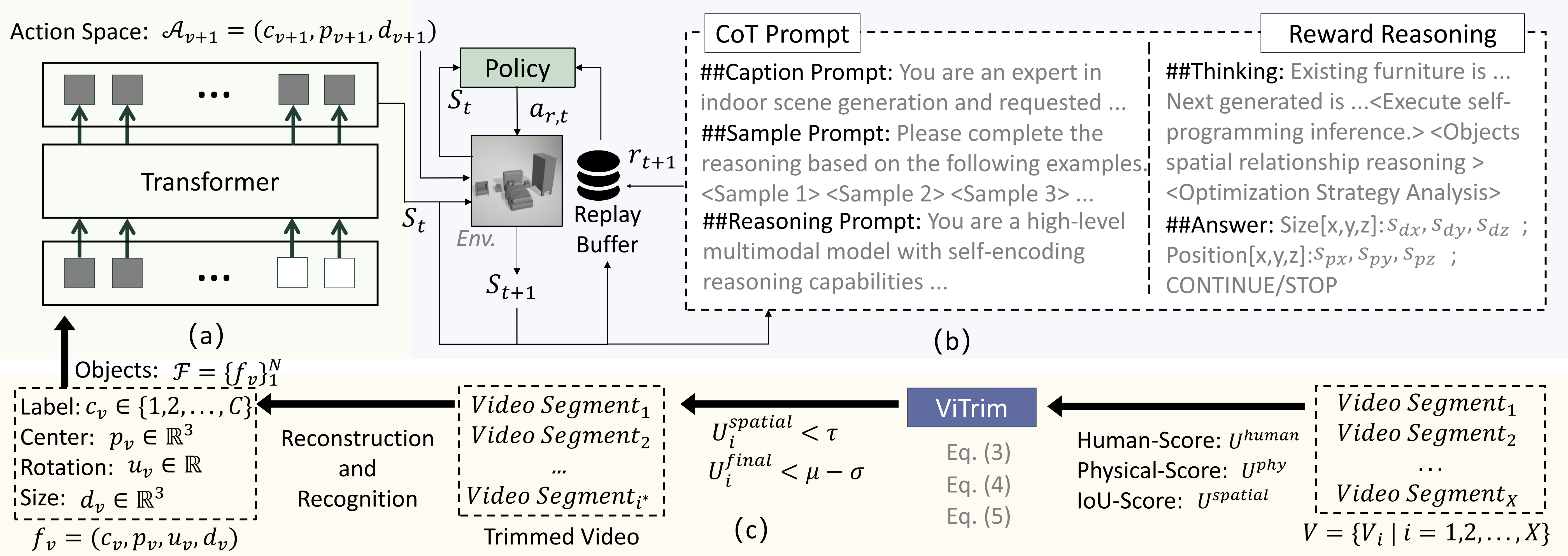}
    \caption{\ours\ pipelines.
    (a) Transformer-based autoregressive 3D indoor scene completion backbone network.
    (b) LLM-driven reinforcement learning fine-tuning loop incorporating CoT spatial reasoning, automatic geometric code execution, and multi-dimensional reward fusion. 
    (c) \trim\ pipeline for high-quality video segment selection.}
    \label{fig:network}
\end{figure*}

\subsection{Pretraining of Generation Backbone}

For training data preparation, we first collect a set of indoor images from the Internet and use them as prompts for a video generation model to produce short video segments. We then apply our \trim\ method to retain only the valid prefix segments from these videos. Following this, we extract 3D objects using a reconstruction and recognition pipeline (e.g., VIPScene~\citep{huang2025video}). The extracted objects are finally recorded to form our training dataset. 

Figure~\ref{fig:network} illustrates the workflow of our approach, which adopts a Transformer backbone. Different from previous work like ATISS~\citep{paschalidou2021atiss}, our training data only contain incomplete scene layouts. Therefore, we make one modification: we remove the learnable end-of-sequence (EOS) token and fix the total generation steps to a predefined constant $N_{\text{max}}$, as shown in Figure~\ref{fig:network}(a). After pre-training, we fine-tune it via reinforcement learning (RL), where the reward signal is derived from an LLM-based reward model.

\subsection{Fine-Tuning via Reinforcement Learning}
\label{sec:rl}

\textbf{Preliminary.}
We model the indoor 3D scene fine-tuning task as a Markov Decision Process (MDP) $(\mathcal{S},\mathcal{A},\mathcal{T},\mathcal{R},\gamma)$, where $s_t \in \mathcal{S}$ is the state space represented by current generated 3D scene parameters, including furniture type, size, and pose; $\mathcal{A}$ is the action space comprising adjustments of object attributes; $\mathcal{T}$ is the state transition function that maps from current state $s_t$ to next state $s_{t+1}$ upon action $a_t$; $\mathcal{R}$ is the reward function generated by our LLM Agent; $\gamma \in [0,1]$ is the discount factor.
At each time step, the agent receives the current state $s_t$ and generates an action $a_t\in\mathcal{A}$ according to the policy $\pi$. Upon executing the action, the environment returns a scalar reward $r_{t+1} \sim \mathcal{R}(S_{t},a_{t})$, the next state $S_{t+1}$, and the done flag.

We adopt the Soft Actor-Critic (SAC) algorithm~\citep{haarnoja2018soft} to alternately optimize the actor network $\pi_\theta$ and critic network $Q_\phi$; the actor and critic loss formulations are provided in Appendix.
Notably, object categories are not part of the action space: they are fixed by the backbone's autoregressive predictions, and the SAC policy refines only the continuous bounding-box attributes $(p_v, u_v, d_v)$ of the current layout.
Generation termination is decided by the LLM's control signal $R^c_t$ rather than by the policy.

\textbf{LLM Reward Design.} The LLM-Agent-based reward mechanism involves three components:

(1) CoT-guided Spatial and Categorical Reasoning. 
Figure~\ref{fig:network}(b) shows the CoT prompt reasoning pipeline.
Given a scene layout, we use role prompting~\citep{john2023art} to guide the LLM to evaluate the category suitability, 3D IoU, and scale matching of the next object $f_{v+1}$ to be generated, assuming existing objects $\mathcal{F} = \{f_v\}_{v=1}^N$. We integrate a prompt-based self-coding mechanism to automatically generate and execute code during reasoning to obtain precise measurements. We further provide manually crafted layout examples and adopt the Chain-of-Thought (CoT) paradigm, which decomposes the complex indoor scene evaluation into explicit, manageable steps covering category verification, size rationality and relative position analysis.

(2) Automatic Code Execution. Following the calculation specifications derived from CoT reasoning, the LLM generates Python code for 3D geometric evaluation. The code executes in an isolated sandbox, taking the full scene state including object categories, 3D coordinates, sizes and rotations as input. It computes quantitative metrics such as 3D bounding box IoU, volume overlap, inter-object distance, floor adherence and category-aware scale matching. All metrics are normalized to $[0,1]$ and directly mapped to position quality score $r_{\text{p}}$ and scale quality score $r_{\text{d}}$.
If code execution fails, LLM iteratively revises the program according to error feedback to maintain stable reward.

(3) Reward Fusion and Early Stopping. We first construct a multi-dimensional reward set $R_t = \{R_{t}^{\text{p}}, R_{t}^{\text{d}}, R_{t}^{\text{c}}\}$ tailored for the CoT reasoning and automatic code execution pipeline. All reward components are normalized to $[0,1]$, with a higher score indicating better scene quality.
$R_{t}^{\text{p}} = \{s_{\text{px}}, s_{\text{py}}, s_{\text{pz}}\}$ and $R_{t}^{\text{d}} = \{s_{\text{dx}}, s_{\text{dy}}, s_{\text{dz}}\}$ denote the position quality and the scale quality scores along the $x$, $y$, and $z$ axes, assessing the distribution of object center positions in terms of inter-object spacing, consistency between each object’s actual size and category-specific statistical priors, as well as functional area compatibility.
Finally, $R_{t}^{\text{c}}=\{0,1\}$ is a discrete control signal derived from the LLM CoT inference output, where ``0" triggers early generation termination.

As shown in Figure~\ref{fig:network}(b), the optimization process follows a closed loop: (1) The actor outputs an initial scene state (with object categories, sizes, poses); (2) LLM performs CoT reasoning and code execution to generate reward $R_t$; (3) SAC uses $R_t$ to update actor/critic networks, adjusting object parameters to obtain a new state; (4) If $R_t$ reaches the predefined threshold, LLM triggers early stopping; otherwise, repeat the loop. When the score gap exceeds a threshold, the LLM revises its reasoning and code to improve evaluation accuracy.

\subsection{3D-Quality Aware Video Trimming (\trim)}
\label{sec:tri}

\textbf{Evaluation Framework.}
 We evaluate video quality using a sliding window. Given a video of total duration $T$, we set window length to 1 s and sliding step to 0.1 s, yielding sampled video segments set $V=\{V_i\mid i=1,2,\dots,X\}$. Each segment is evaluated from three complementary dimensions, with all scores normalized to \([0,1]\): (1) VisionReward score~\citep{xu2024visionreward}~$U^\text{human}$, which quantifies visual naturalness and human preference of indoor scenes; (2) VideoPhy score~\citep{bansal2024videophy}~$U^\text{phy}$, which measures whether object arrangements comply with physical constraints; (3) Average object IoU~$U^\text{spatial}$, which is proposed in this work to reflect spatial validity by penalizing redundant overlap between object pairs. The final evaluation score is an entropy-based weighted~\citep{shannon1948mathematical,kiker2005application} sum of these scores, penalized by cumulative spatial error. The evaluation procedure is illustrated in Figure~\ref{fig:network}(c).

\textbf{Entropy-based Score Fusion.}
\label{sec:fusionscore}
Given a metric space $U=\{U^j \mid j=1,2,\dots,Y\}$, where $U^j$ represents the $j$-th metric. For the $j$-th metric, its information entropy $E^j$ is:
\begin{equation}
E^j = -\frac{1}{\ln X} \sum_{i=1}^X p_{i}^{j} \ln\left(p_{i}^{j} + \epsilon\right),
\end{equation}
where $p_{i}^{j} = U_{i}^{j}/{\sum_{i=1}^X U_{i}}^{j}$ is the normalized proportion of the $i$-th video segment, and $\epsilon = 10^{-8}$ avoids logarithmic singularity.
When the denominator $\sum_{i=1}^X U_{i}^{j} = 0$, we uniformly set $p_{i}^{j} = 1/X$ for all video segments to ensure numerical stability.
A lower value of $E^j$ implies a stronger discriminative degree $D^j = 1 - E^j$.
We assign weights to the metric $w^j = {D^j}/{\sum_{j=1}^Y D^j}$, where $\sum_{j=1}^Y w^j = 1$.
With the calculated objective weights, we define the fusion score for each video segment as:
\begin{equation}
U^{\text{fused}}_{i} = w^{\text{human}} \cdot U^{\text{human}}_{i} + w^{\text{phy}} \cdot U^{\text{phy}}_{i} + w^{\text{spatial}} \cdot U^{\text{spatial}}_{i}.
\end{equation}

\textbf{Cumulative Spatial Error as Penalty.}
\label{sec:Temporal}
Per-segment IoU only measures short-term video quality and does not reflect the progressive deterioration of scene quality commonly found in generated videos.
To capture such an effect, we introduce cumulative spatial error. Given a set of spatial IoU scores $U^\text{spatial}$,
we define the cumulative spatial error, for the video segments from 1 to $i$, as~\(E^{\text{cum}}_{i} = \sum_{k=2}^i  (U^{\text{spatial}}_{1}-U^{\text{spatial}}_{k})\).
We take the first video segment as a baseline, which is often of high quality but can exhibit certain spatial overlap due to the bounding box approximations, and add up the spatial error of each segment relative to the first one. We normalize this error term to $[0,1]$, and incorporate it as a penalty into the fused score, obtaining the final validity score:
\begin{equation}
U^{\text{final}}_{i} = \text{segment}\left( U^{\text{fused}}_{i} - \gamma \cdot E_{i}^{cum},\ 0,\ 1 \right),
\end{equation}
where $\gamma>0$ controls penalty strength ($\gamma=0.5$ in our experiments), and $\text{segment}(\cdot)$ constrains scores within $[0,1]$.

\textbf{Hard-Soft Constraints Guided Video Trimming.}
\trim\ applies both hard and soft constraints to select valid video segments.
A segment satisfying $U^{\text{spatial}}_{i} < \tau$ is immediately rejected, where $\tau$ is a hard spatial threshold ($\tau=0.5$ in our experiments), as severe spatial overlap clearly violates physical plausibility. Also, we compute the global mean $\mu$ and standard deviation $\sigma$ of all $U^{\text{final}}_{i}$ within a video. We set the soft quality threshold as $T_\sigma = \mu + \beta \cdot \sigma$, where $\beta$ is the weight of $\sigma$, and define the {\em trimming point} $i^\star$ as the first segment that violates either constraint: $i^\star = \min\{ i \mid U^{\text{spatial}}_{i} < \tau \ \text{or}\ U^{\text{final}}_{i} < T_\sigma \}$, where $U^{\text{spatial}}_{i}=1-IoU_i$. All segments preceding $i^\star$ are retained as the valid clip: $\mathcal{V}_{\text{valid}} = \{ V_1, V_2, \dots, V_{i^\star-1} \}$, which is passed to the 3D reconstruction pipeline.

\begin{table*}[t]
\centering
\setlength{\abovecaptionskip}{10pt}
\setlength{\belowcaptionskip}{5pt}
\resizebox{\textwidth}{!}{%
    \begin{tabular}{l ccccc ccccc}
        \toprule
        \multirow{2}{*}{\textbf{Method}}  & \multicolumn{5}{c}{\textbf{Bedroom}} & \multicolumn{5}{c}{\textbf{LivingDiningRoom}}\\
        \cmidrule(lr){2-6} \cmidrule(lr){7-11}
        & FID~$\downarrow$ & KID~$\downarrow$ & SCA~\% & CKL~$\downarrow$ & PIoU~$\downarrow$ 
        & FID~$\downarrow$ & KID~$\downarrow$ & SCA~\% & CKL~$\downarrow$ & PIoU~$\downarrow$ \\
        \midrule
        ATISS & 
        123.06 & 0.70 & 57.20 & 1.85 & 0.48 & 135.27 & 7.98 & 67.72 & 2.78 & 1.52 \\
        DiffuScene & 
        124.19 & 0.62 & 60.57 & 1.59 & 0.46 & 141.45 & 7.75 & 70.50 & 1.56 & 0.69 \\
        VIPScene & 
        137.81 & 0.67 & 68.01 & 2.19 & 0.36 & 147.36 & 9.05 & 74.32 & 2.59 & 0.34 \\
        Ours & 
        \textbf{111.06} & \textbf{0.56} & \textbf{55.65} & \textbf{1.13} & \textbf{0.25} & \textbf{128.57} & \textbf{5.64} & \textbf{62.58} & \textbf{1.38} & \textbf{0.27} \\
        \bottomrule
    \end{tabular}%
}
\caption{Quantitative results. ATISS~\citep{paschalidou2021atiss}, DiffuScene~\citep{tang2024diffuscene}, and VIPScene~\citep{huang2025video} are compared. \ours\ performs best across all metrics. PIoU ($\downarrow$) is newly introduced to measure physical plausibility; baseline PIoU values are computed on their released scenes.}
\label{tab:combined_scores}
\end{table*}

\begin{figure*}
  \centering
     \includegraphics[width=\linewidth]{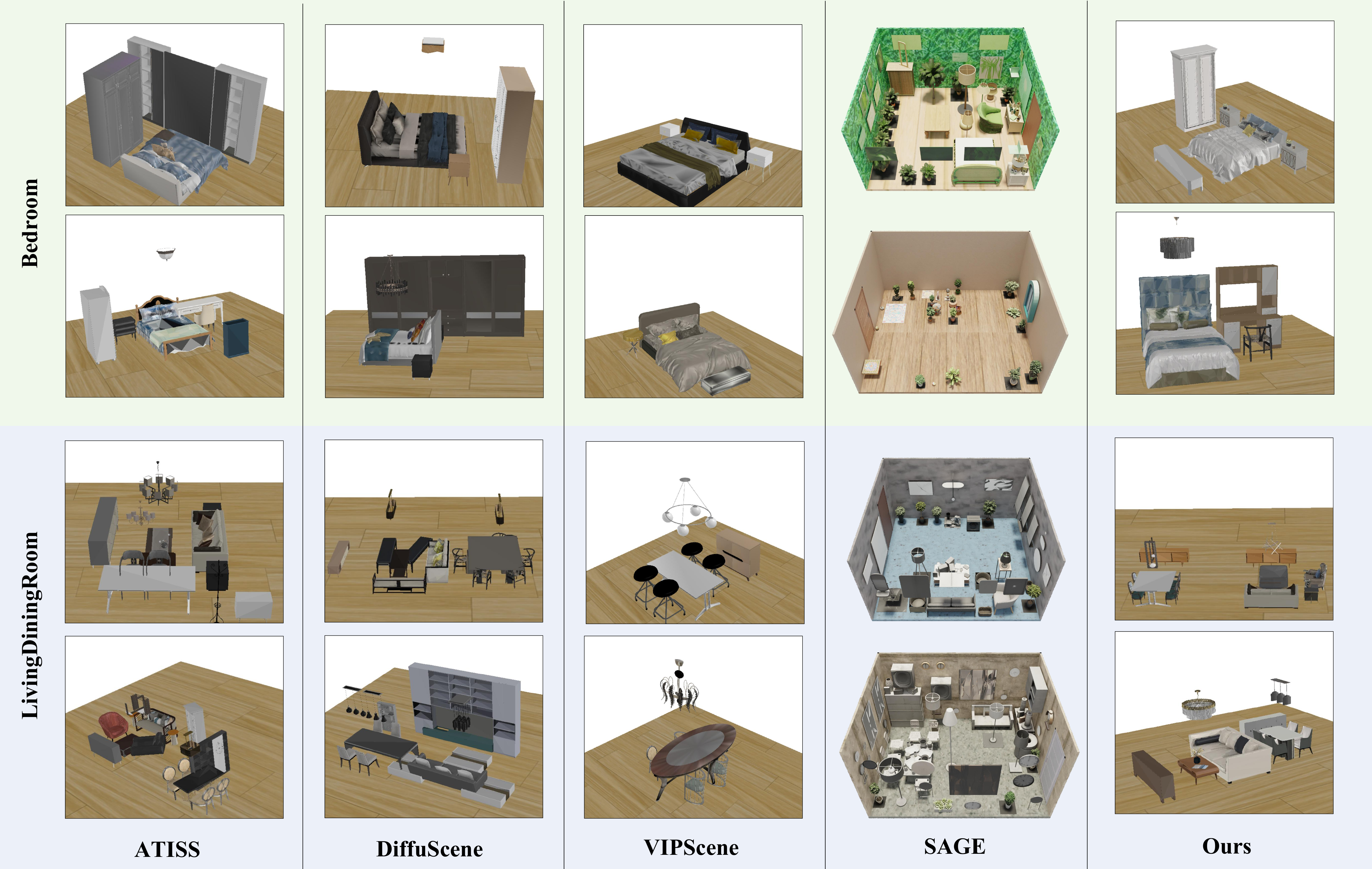}
     \caption{Generation samples from an empty scene. \ours\ performs well in producing reasonable and complete object layouts, with correct collision avoidance.
     ATISS and DiffuScene generate misaligned objects and unnatural layouts.
     Perhaps restricted by the video generation model, VIPScene generates a small number of objects with narrow coverage of the scene, missing objects that are commonly found in realistic scenes (e.g., wardrobe for bedroom, sofa for livingdiningroom).
     SAGE~\citep{xia2026sage} lacks the understanding of the actual physical space, resulting in the generation of duplicate or poorly arranged objects (e.g., bedroom without bed, impassable passageway).}
     \label{fig:result1}
\end{figure*}

\begin{figure*}
  \centering
     \includegraphics[width=0.9\linewidth]{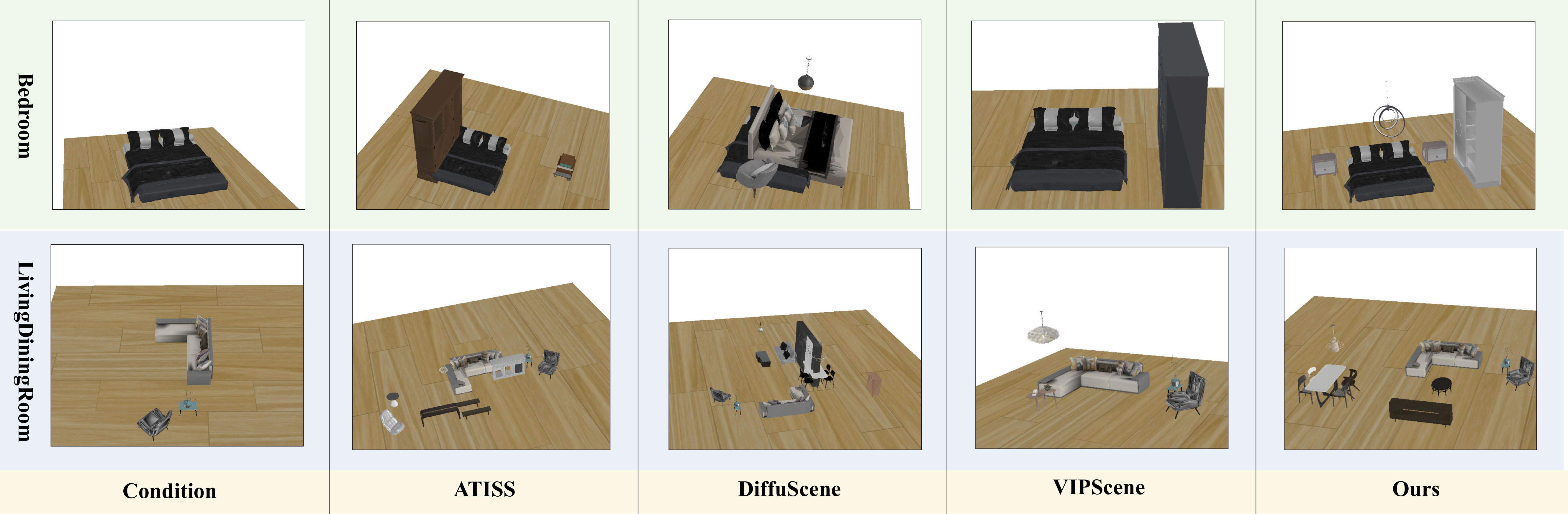}
     \caption{Scene completion results from partial scenes. Starting from a few objects, \ours\ completes the scenes more thoroughly and plausibly.}
     \label{fig:result1_2}
\end{figure*}

\section{Experiments}

\noindent\textbf{Datasets and Experimental Details.}
\label{sec:Datasets and Experimental details}
We collected 20,000 bedroom images and 20,000 living room images from Google and the image collection of 3D-FRONT (see Appendix for details).
Each image serves as the initial frame for LTX-Video~\citep{hacohen2024ltx}, which synthesizes a 5-second video that preserves the original scene layout with temporally consistent frames. Following VIPScene~\citep{huang2025video}, we apply Grounded-SAM~\citep{ren2024grounded} for object segmentation, UniDepth~\citep{piccinelli2024unidepth} for depth estimation, MASt3R~\citep{leroy2024grounding} for cross-frame correspondence, and Fast3R~\citep{yang2025fast3r} for 3D scene reconstruction. We evaluate all methods on living room, dining room, and bedroom scenes.
Our model is trained on no 3D layout annotation data and rendered using 3D assets from 3D-FRONT~\citep{fu20213d} (see Appendix for a discussion of generalization).
Specifically, we populate each predicted box $d_v$ by selecting the 3D-FUTURE asset whose normalized dimension vector $(w/h, d/h)$ minimizes the $\ell_1$ distance to that of $d_v$, so that the rendered object fits the predicted envelope without rescaling or deformation.
Further experimental details and hyperparameters are provided in Appendix.

\noindent\textbf{Metrics.}
\label{sec:Metrics}
Following prior work~\citep{paschalidou2021atiss, tang2024diffuscene}, we report FID, KID ($\times 0.001$), SCA, CKL ($\times 0.01$), and PIoU ($\times 0.01$). Lower FID, KID, CKL, and PIoU indicate better fidelity, diversity, category alignment, and spatial plausibility, respectively; an SCA near 50\% indicates generated scenes are indistinguishable from real ones. Full metric definitions are provided in Appendix.

\noindent\textbf{Baselines.}
We compare against ATISS~\citep{paschalidou2021atiss}, DiffuScene~\citep{tang2024diffuscene}, and VIPScene~\citep{huang2025video}. ATISS and DiffuScene are both trained on annotated 3D-FRONT~\citep{fu20213d} data: ATISS generates scenes autoregressively over a discretized object space, while DiffuScene uses a diffusion model. VIPScene reconstructs 3D scenes from video by drawing on physical commonsense encoded in a video generation model, and requires no annotated data. Since VIPScene provides no per-scene ground truth, we follow standard protocols and evaluate all methods on global distribution metrics, ensuring a fair comparison.
We also compare the generation quality against the recent agent-based method SAGE~\citep{xia2026sage}.

\begin{figure*}
  \centering
     \includegraphics[width=0.91\linewidth]{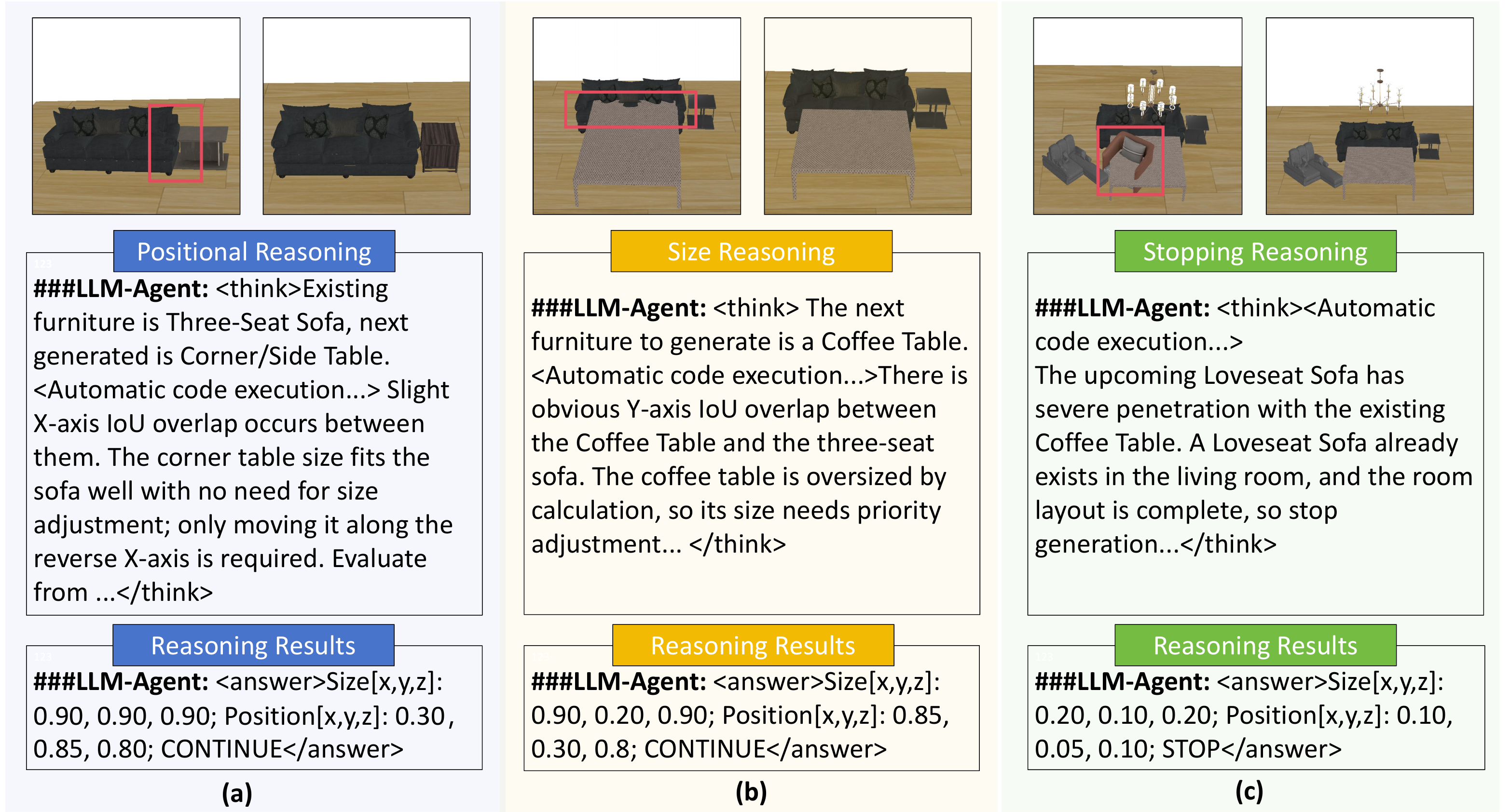}
     \caption{LLM reasoning process.
     (a) Position refinement for minor spatial overlap; 
     (b) Size correction for severe object interpenetration; 
     (c) Layout saturation detection and generation stopping.
     The LLM Agent applies CoT reasoning and self-programmed geometric computation. 
     In each case, Size[x,y,z] denotes the corrected 3D dimension of an object, Position[x,y,z] its refined 3D centroid. CONTINUE indicates proceeding with object generation, while STOP terminates the process when layout is complete or spatially invalid.}
     \label{fig:result4}
\end{figure*}

\subsection{Experimental Results}
\label{sec:experiment}

Table~\ref{tab:combined_scores} reports quantitative results for all methods.
Although SAGE tackles a similar task, it does not learn from real scene data, so it cannot be fairly compared on 3D-FRONT metrics and scene completion.
For SAGE, we report qualitative results only.
Our approach outperforms all supervised and video-based baselines across metrics. Specifically, compared to DiffuScene, SceneCast improves FID by 10\% and KID by 18\%, demonstrating stronger alignment with the real scene distribution and greater diversity. Our SCA score is closest to 50\%, confirming that the generated scenes are nearly indistinguishable from real ones. Critically, SceneCast achieves the lowest PIoU, indicating substantially fewer inter-object collisions and thus more physically plausible layouts---a dimension on which the distributional metrics (FID/KID) are silent. ATISS and DiffuScene require costly labeled training data, whereas SceneCast is annotation-free. VIPScene also requires no labels, yet underperforms across the metrics, because it reconstructs only the 3D scene visible in the partial generated video and never observes the full scene distribution.

Figure~\ref{fig:result1} visualizes scene generation results from an empty scene. 
The agent-based SAGE could produce more complete scenes; however, due to a lack of real-world scene distributions, the generated objects often exhibit redundancy and unreasonable spatial arrangements.
Although ATISS and DiffuScene produce generally reasonable layouts, they frequently exhibit object misalignment and interpenetration, yielding physically implausible arrangements. Both methods rely solely on statistical patterns in the training data and receive no spatial commonsense guidance.
When the learned distribution fails to cover certain cases, the models produce layouts that violate physical plausibility.
Figure~\ref{fig:result1_2} shows the results of scene completion starting from a few objects, a set of experiments revealing more details of how generated objects interact with existing ones.  
Due to the lack of explicit constraints, ATISS and DiffuScene can produce object categories that conflict with existing parts of the scene during scene completion (e.g., producing an extra sofa when a sofa already exists, or producing an extra bed when one already exists.).
While VIPScene can usually reconstruct a reasonable layout from a video, the number and types of objects it generates are limited, resulting in an incomplete layout.
In contrast, \ours\ generates complete, physically plausible layouts by drawing on the spatial commonsense encoded in the LLM reward.

\subsection{Ablation study}
\label{sec:Ablation study}

\noindent\textbf{LLM Reward Mechanism.}
Figure~\ref{fig:result4} presents explicit CoT reasoning traces and numerical outputs from the LLM reward mechanism. In Figure~\ref{fig:result4}(a), the LLM reasons over the spatial relationship between an existing three-seat sofa and a newly generated corner table. Through automatic code execution, it computes 3D IoU and detects a slight overlap along the X-axis. Since the object sizes are already within acceptable bounds, the LLM refines the position parameters to resolve the overlap and returns a \textsc{Continue} signal to proceed with generation. In Figure~\ref{fig:result4}(b), faced with severe interpenetration, the LLM prioritizes size calibration to eliminate the conflict. In Figure~\ref{fig:result4}(c), foreseeing simultaneous penetration and redundant object placement, the LLM determines that the scene is complete and issues an early stopping signal.
These examples show that the LLM reward captures spatial constraints (object conflicts, scale errors, redundant generation) that fixed rules cannot easily encode. We verify this quantitatively below.

\noindent\textbf{Reward Design Ablation.}
We isolate the contribution of each component in the reward design by progressively augmenting the reward signal, keeping the SAC pipeline identical across all variants. Starting from the pre-RL backbone (\textbf{No RL}), we first add a purely algorithmic geometric reward (\textbf{Rule-Geo}) that penalizes collisions, enforces category-specific size ranges, and checks floor contact. We then augment it with hand-crafted categorical rules (\textbf{Rule+Cat}), such as ``at most one bed per bedroom'' and ``nightstands adjacent to the bed.'' Next, we replace the rules with the LLM reward but disable code execution (\textbf{LLM w/o code}), so that the LLM reasons about geometry in natural language only. Finally, we enable automatic code execution (\textbf{LLM w/ code}), our full configuration. Table~\ref{tab:ablations}~(a) reports the results, where each row differs from the one above by exactly one component.

First, geometric reinforcement learning already improves over the backbone, but hand-crafted rules (Rule+Cat) yield only marginal gains, as a fixed rule set cannot enumerate the diversity of functional arrangements.
Second, replacing rules with LLM reasoning (LLM w/o code) brings a substantial improvement, showing that open-ended spatial reasoning matters more than enumerated rules.
Third, enabling code execution further reduces PIoU by 0.27, showing that precise geometric computation, not language reasoning alone, is essential for reliable reward. Together, these results justify both the use of an LLM and the code-execution mechanism.

\begin{table}[t]
\centering
\footnotesize
\setlength{\tabcolsep}{3pt}
\begin{minipage}[t]{0.31\linewidth}
\centering
\begin{tabular}{@{}lcc@{}}
\toprule
\textbf{Reward config.} & \textbf{FID}~$\downarrow$ & \textbf{PIoU}~$\downarrow$ \\
\midrule
No RL & 156.6 & 1.34 \\
Rule-Geo & 147.3 & 0.78 \\
Rule+Cat & 143.5 & 0.72 \\
LLM w/o code & 141.3 & 0.52 \\
LLM w/ code & \textbf{111.1} & \textbf{0.25} \\
\bottomrule
\end{tabular}

\vspace{3pt}
(a) Reward ablation (bedroom)
\end{minipage}
\hfill
\begin{minipage}[t]{0.31\linewidth}
\centering
\begin{tabular}{@{}lcc@{}}
\toprule
\textbf{Curation strategy} & \textbf{FID}~$\downarrow$ & \textbf{KID}~$\downarrow$ \\
\midrule
No filtering & 133.9 & 0.69 \\
Fixed 1.5\,s prefix & 136.2 & 0.70 \\
VR+VP only & 132.2 & 0.61 \\
\trim\ (equal $w$) & 127.1 & 0.64 \\
\trim\ (entropy) & \textbf{111.1} & \textbf{0.56} \\
\bottomrule
\end{tabular}

\vspace{3pt}
(b) ViTrim ablation (bedroom)
\end{minipage}
\hfill
\begin{minipage}[t]{0.31\linewidth}
\centering
\begin{tabular}{@{}lcc@{}}
\toprule
\textbf{Mask ratio} & \textbf{HCR}~$\uparrow$ & \textbf{Dup}~$\downarrow$ \\
\midrule
25\% & 0.67 & 0.18 \\
50\% & 0.58 & 0.17 \\
75\% & 0.34 & 0.11 \\
\bottomrule
\end{tabular}

\vspace{3pt}
(c) Conditional completion
\end{minipage}
\caption{Ablations on the bedroom set. (a) Unified reward ablation. Each row augments the previous one with a single component, isolating its effect. The LLM reward outperforms hand-crafted rules, and code execution provides a further gain over language-only reasoning.
(b) \trim\ ablation. All models share identical architecture and training hyperparameters, while only the data curation strategy differs. Rows 1–2 isolate the effect of trimming, row 3 isolates the spatial IoU signal and rows 4–5 compare entropy-based vs uniform weighting.(c) \ours\ completion quality across condition coverage on held-out 3D-FRONT scenes.}
\label{tab:ablations}
\end{table}

\noindent\textbf{Effect of \trim.}
\trim\ fuses two quality metrics (VisionReward, VideoPhy) with a new spatial IoU signal via entropy-based weighting. To isolate each factor, we train five otherwise identical models on the bedroom set, varying only the data curation strategy.
Table~\ref{tab:ablations}~(b) reports the results. Comparing row~1 with row~5, discarding corrupt video suffixes reduces FID by 22.8, confirming that untrimmed videos introduce noisy training samples.
The fixed 1.5 s prefix (row 2) performs no better than no filtering (133.9), suggesting that a fixed cut-off discards useful content in some videos while retaining corruption in others.
Comparing row~3 with row~5, adding the spatial IoU signal further reduces FID by 21.1, showing that perceptual and physical quality scores alone are insufficient to detect geometric interpenetration. Comparing row~4 with row~5, entropy-based weighting adapts the cut-off to each video's quality profile and outperforms a uniform $\tfrac{1}{3}$ allocation. Qualitative visualization is provided in Appendix.

\textbf{Conditional completion analysis.}
Following the partial-scene completion protocol of DiffuScene~\citep{tang2024diffuscene}, we mask 25\%, 50\%, and 75\% of the objects in scenes and evaluate the result with category-based metrics.
We adapt the category recall convention: hidden-category recall (HCR) is the fraction of masked categories regenerated, and duplicate rate (Dup) is the fraction of generated objects duplicating an input category beyond its input count. Instance-level matching is ill-posed for generative models, hence both metrics are category-based. Table~\ref{tab:ablations}~(c) evaluates \ours's conditional completion across condition coverage. HCR decreases monotonically as more objects are masked, confirming that completion is condition-driven; Dup remains low across all ratios, indicating that duplicating an existing category is rare for \ours.
\section{Conclusion}
Indoor scene generation has long been constrained by the cost of annotated training data. This work trains a scene completion model on unlabeled video data through two core designs: a Transformer backbone fine-tuned with LLM-guided reinforcement learning to suppress extrapolation errors and enforce spatial commonsense, and a 3D quality-aware video trimming strategy that selects synthetic training clips through multi-criteria scoring and filtering. On the 3D-FRONT benchmark, our method outperforms fully supervised baselines. Extending the model with language conditioning to enable text-to-scene generation is a natural next step, which we leave for future work. 

\clearpage
\subsection*{AI use statement}
Video generation models were used to construct the dataset, and generative AI tools were used during the preparation of this work to assist with language editing and improving the clarity and organization of the manuscript, as well as for literature search and summarization. All research ideas, methodological decisions, experimental results, and conclusions were independently examined and validated by the authors. Relevant literature suggested by AI tools was manually verified against the original sources, and all AI-assisted text and technical content were reviewed and revised by the authors. The authors take full responsibility for the final content of this work, including all text, claims, results, and artifacts produced with the aid of generative AI.

{
    \small
    \bibliographystyle{ieeenat_fullname}
    \bibliography{main}
}

\end{document}